\documentclass[letterpaper]{article} % DO NOT CHANGE THIS
\usepackage{aaai24}  % DO NOT CHANGE THIS
\usepackage{times}  % DO NOT CHANGE THIS
\usepackage{helvet}  % DO NOT CHANGE THIS
\usepackage{courier}  % DO NOT CHANGE THIS
\usepackage[hyphens]{url}  % DO NOT CHANGE THIS
\usepackage{graphicx} % DO NOT CHANGE THIS
\usepackage{natbib}  % DO NOT CHANGE THIS AND DO NOT ADD ANY OPTIONS TO IT
\usepackage{caption} % DO NOT CHANGE THIS AND DO NOT ADD ANY OPTIONS TO IT
\usepackage{lipsum}
\usepackage{algorithm}
\usepackage{algorithmic}
\usepackage{xcolor}
\usepackage{float}
\usepackage{subcaption}
\usepackage{newfloat}
\usepackage{listings}
\usepackage{float}
\DeclareCaptionStyle{ruled}{labelfont=normalfont,labelsep=colon,strut=off} % DO NOT CHANGE THIS
\floatstyle{ruled}
\newfloat{listing}{tb}{lst}{}
\floatname{listing}{Listing}
\title{LLMGuard: Guarding against Unsafe LLM Behavior }
\author {
Shubh Goyal \textsuperscript{\rm1}\thanks{Equal Contribution},
Medha Hira\textsuperscript{\rm2}\footnotemark[1],
Shubham Mishra \textsuperscript{\rm1}\footnotemark[1],
Sukriti Goyal \textsuperscript{\rm1}\footnotemark[1],
Arnav Goel\textsuperscript{\rm2}\footnotemark[1],
Niharika Dadu \textsuperscript{\rm1}\footnotemark[1],
Kirushikesh DB \textsuperscript{\rm3},
Sameep Mehta \textsuperscript{\rm3},
Nishtha Madaan\textsuperscript{\rm3}
}
\affiliations {
    \textsuperscript{\rm 1}Indian Institute of Technology (IIT), Jodhpur, India \\ 
    \textsuperscript{\rm 2} Indraprastha Institute of Information Technology Delhi (IIIT-D)\\
    \textsuperscript{\rm 3} IBM Research, India\\
}

\usepackage{bibentry}
\begin{document}

\maketitle

\begin{abstract}
Although the rise of Large Language Models (LLMs) in enterprise settings brings new opportunities and capabilities, it also brings challenges, such as the risk of generating inappropriate, biased, or misleading content that violates regulations and can have legal concerns. \footnote{\url{https://www.tcs.com/what-we-do/pace-innovation/article/generative-ai-guardrails-secure-llm-usage}}.
% Guardrails are essential to draw a boundary across the generative capability of LLMs while maintaining the foundational trust in AI. 
To alleviate this, we present \emph{"LLMGuard"}, a tool that monitors user interactions with an LLM application and flags content against specific behaviours or conversation topics. To do this robustly, LLMGuard employs an ensemble of detectors.

% a system study presents \textbf{Five Guardrails}: Personal Identifiable Information (PII) detection, Bias detection, Toxicity/Hate-Speech detection, Violence Detection and Blacklisted Topics detection. The detectors monitor user interactions with an LLM application and flag the content against specific behaviours or conversation topics.
\end{abstract}

\section{Introduction and Related Work}
% LLM's based applications have become part of most of the organisations today. While building internal tools and client facing tools, organisations re-use existing LLM's.   

% \begin{figure*}[h!]
% \centering
%   \includegraphics[width=1.0\linewidth]{Demo_Paper_Diagram.png}
%   \caption{Architecture of \emph{LLMGuard}. The user input and the LLM response are provided to an ensemble of 5 detectors. If any detectors flag the text as unsafe, the transaction is blocked.} 
% \label{fig:arch}
% \end{figure*}

% \textcolor{red}{\lipsum[1]}

% \textcolor{red}{\lipsum[1]}

% \textcolor{red}{\lipsum[1]}

Large Language Models (LLMs) have risen in importance due to their remarkable performance across various NLP tasks, including text generation, translation, summarization, question-answering, and sentiment analysis \cite{muneer2020comparative, goel2023advancements, kalyan2021ammus}. LLMs serve as a general-purpose language task solver to some extent, and the research paradigm has been shifting towards using them \cite{zhao2023survey}. With the advent of models such as PaLM \cite{chowdhery2022palm}, GPT-3 \cite{brown2020language} and GPT-4 \cite{openai2023gpt4}, LLMs have found increased use-cases in domains such as the medicine \cite{kitamura2023chatgpt}, education \cite{peng2021mathbert}, finance and entertainment \cite{dowling2023chatgpt}. 
% Place Holder for Figure 1

% Despite their phenomenal success, state-of-the-art LLMs still lack many aspects. 

% This makes them less likely as a case for the early manifestation of AGI \cite{hadi2023large}. 
% Large Language Models (LLMs) require a large corpus of data for pre-training the model. Collecting and curating these datasets can be highly challenging. 

Despite their phenomenal success, LLMs often exhibit behaviours that make them unsafe in various enterprise settings. For instance, the text can contain confidential or personal information, such as telephone numbers, leading to privacy leaks \cite{kaddour2023minipile}. Instances of bias have also been reported in LLM responses, raising ethical concerns when deploying them in various applications\cite{kaddour2023challenges}. \cite{viswanath2023fairpy} presents a comprehensive quantitative evaluation of different kinds of biases, such as race, gender, ethnicity, age, etc., exhibited by recent LLMs. Such risks raise concerns about the implications of the growing use of LLMs in different areas, from education to heritage to healthcare \cite{urman2023silence}. 

% As enterprises and startups have begun providing services powered by LLMs\footnote{\url{https://www.ibm.com/watsonx}}, it becomes essential for organizations to ensure their reliability. 

To address them, various techniques have been proposed to align LLMs with human preferences, such as RLHF, which finetune the model based on safety and helpfulness objectives \cite{touvron2023llama}. Another approach focuses on red flagging and rectifying undesirable language behaviours \cite{perez2022red}. However, constant retraining is necessary in such techniques, making them prohibitive in many cases. A promising line of methods pursues post-processing to apply guardrails directly to the LLM outputs. This ensures they stay within specific parameters by validating user and LLM responses. 

\begin{figure*}[ht!]
\centering
  \includegraphics[width=1\linewidth]{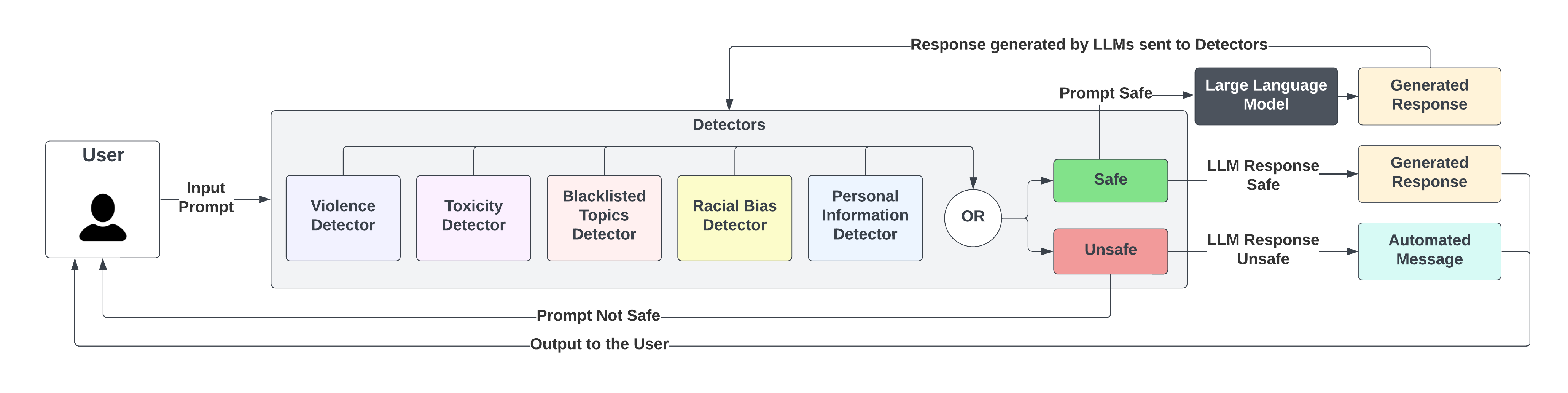}
  \caption{Architecture of \emph{LLMGuard}. The user input and the LLM response are provided to an ensemble of 5 detectors. If any detectors flag the text as unsafe, the transaction is blocked.} 
\label{fig:arch}
\end{figure*}

In this work, we propose a tool \emph{LLMGuard}, which employs a library of detectors to post-process user questions and LLM responses. These detectors help flag undesirable inputs and responses such as Personal Identifiable Information (PII), bias, toxicity, violence, and blacklisted topics. Lastly, we provide a demo of how LLMGuard works on two recent LLMs: FLAN-T5 and GPT-2 \cite{chung2022scaling, radford2019language}, and show the effectiveness of our framework. A high-level architecture of our tool is shown in Figure \ref{fig:arch}.\\

% We implemented five guardrails for detecting \textbf{Personal Identifiable Information (PII), Bias, Toxicity, Violence, and Blacklisted Topics}. This integration with LLM chatbots prevents engagement with biased or inappropriate content.

% Various techniques have been proposed to align LLMs with human preference, such as Reinforcement Learning with Human Feedback(RLHF), which tunes the model based on safety and helpfulness \cite {touvron2023llama}. The other approach around Language Mode (LM) based red teaming is a promising tool for finding and fixing diverse, undesirable LM behaviours before impacting users\cite{perez2022red}. We believe, given the emerging recognition of risks associated with LLM, it is no longer practical to repeatedly retrain these models. One viable solution is to implement guardrails as a post-processing step. The goal of guardrails is to enforce the output of an LLM to be in a specific format or context while validating each response from the user and the LLM. In our work, we implemented five guardrails for handling different types of cases ranging from Personal Identifiable Information (PII) detection, Bias detection, Toxicity detection, Violence detection and Blacklisted Topics detection. By integrating guardrails with LLM chatbots, we can suppress LLM from engaging with biased content or specific discussions on unwanted topics.
\vspace{-2mm}

\section{Method: LLMGuard}
This section describes our proposed tool called \textit{LLMGuard} for detecting and preventing undesirable LLM behaviour.  
% When a user inputs its prompt, it is sent to the Detectors' Module, which parallelly checks for it against the five guardrails. If the prompt does not raise an alarm with any of the detectors, it is sent to any black-box LLM. The response generated is then sent back to the Detectors' Module. If it raises an alarm in any one detector, an automated message is sent back to the user instead of the LLM-generated response. Implementation of the five guardrails is described as follows
Broadly, LLMGuard works by passing every user prompt and every LLM response via an ensemble of detectors. If any of the detectors detect unsafe text, an automated message is sent back to the user instead of the LLM-generated response. 
We now describe the detectors we employ in our ensemble.

\subsection{Library of Detectors}
In LLMGuard, the ensemble consists of a library of detectors. It provides a modular framework for easily adding, modifying or removing the detectors within the ensemble. Each detector is an expert in detecting a specific unsafe behavior and operates independently of the other detectors. We now describe each of our 5 detectors in detail.

\subsubsection{Racial Bias Detector}
This detector seeks to flag prejudiced or discriminatory content towards a particular race or community. 
% It poses a significant challenge as this bias can manifest in various ways, from explicit discrimination to subtle, implicit prejudices. 
We implement the detector using an LSTM \cite{hochreiter1997long}. 
% To implement this detector, we first map the given input text to a sequence of word embeddings. We then provide these embeddings to an LSTM model with a hidden size of 100. The final output is fed to an MLP, followed by Softmax, to predict the probability that racial bias exists in the input text. For stability, we employed a dropout of 0.2 during training. 
The detector was trained on the Twitter Texts Dataset \cite{go2009twitter} comprising 27500 tweets. The detector obtains an accuracy of 87.2\% and an F1 score of 85.47\% on the test set.

% Our proposed method consists of a dense-embedding layer followed by \textbf{Spatial Dropout} layer with a dropout rate 0.2. 
% % Spatial dropout is an effective regularization technique that helps prevent overfitting by randomly dropping entire 1D feature maps. 
% This is followed by a 100-neuron LSTM layer with a 0.2 recurrent dropout. The outputs from the LSTM are sent through a Dense Layer, and probabilities are calculated using Softmax.

% We used Categorical Cross Entropy Loss for training our model on the Twitter Texts Dataset comprising 27500 tweets. Our model obtained an accuracy of 87.2\% and a F1 score of 85. 47\% on the testing set.

\subsubsection{Violence Detector}
This detector seeks to flag the presence of threats and violence in an LLM-generated response. To implement this detector, we employ a simple count-based mapping to vectorise our text. An MLP is followed by a sigmoid layer to predict the probability of the presence of violence or threat in the text. The model was trained on the Jigsaw Toxicity Dataset 2021 \cite{10.1145/3038912.3052591} and achieved an accuracy of 86.4\%.
% \textcolor{red}{\lipsum[1]}

\subsubsection{Blacklisted Topics}
This detector seeks to flag the presence of sensitive or blacklisted topics. What topics to blacklist is provided by the user in a plug-and-play manner. 
% For example, if an enterprise considers information about its competitors' products undesirable, it may blacklist content about its competitors. 
In our current version, we consider \textit{Politics}, \textit{Religion} and \textit{Sports} as blacklisted categories. To implement this detector, we fine-tune a BERT model \cite{devlin-etal-2019-bert} on the 20-NewsGroup Dataset \cite{misc_twenty_newsgroups_113} containing text about politics, religion and sports and their topic labels. The classifier for each blacklisted topic is independently trained such that one may easily enable or disable a certain topic. Our detector achieves an average accuracy of $\approx$92\% for the classifiers corresponding to these topics.

% Blacklisted Topic Detection refers to detecting the presence of topics in the interactions between the users and the LLM, which are blacklisted by the users and filter those out with a standard response. The objective is to customize the list of topics, allowing users to add their topics with sample data and prevent the application from responding to those topics. For example, an enterprise doesn't want its chatbot to respond with information about its competitors' products. 

% The recent work involving Nvidia's Nemo Guardrails\footnote{\url{https://blogs.nvidia.com/blog/2023/04/25/ai-chatbot-guardrails-nemo/}} uses semantic search to identify if the text falls into the blacklisted topic and performs the corresponding action. In our study, we considered taboo topics such as religion, politics, and sports as blacklisted, and we used bert-based-uncased as a base model for fine-tuning each topic using the 1v1 approach. During the inference, if any detectors detect text as blacklisted with a score of 0.5 or higher, they are mapped as blacklisted. The combined system achieved an accuracy and F1 score of 79.6\% and 78.0\% respectively on the Sklearn-20 NewsGroup Dataset.

\subsubsection{PII Detector}
The detector seeks to flag Personal Identifiable Information (PII). Users often provide sensitive information to LLM, such as names, addresses, emails, IP addresses and phone numbers. We detect such content through regular expressions to identify specific PII and ensure that such information is not shared with the LLM. Our model achieves an NER F1-score of 85\%.
% Personal Identifiable Information (PII) detection is essential for data security and privacy preservation. PII encompasses a wide range of sensitive information, including names, addresses, email addresses, IP addresses, and phone numbers. We have developed a PII detection and anonymization mechanism. Our method uses spaCy's 'en\_core\_web\_sm'\footnote{\url{https://spacy.io/models/en#en_core_web_sm}} NER model, pre-trained on web text like blogs and news articles. Furthermore, we employ regular expression (regex) patterns to identify and anonymize specific PII elements such as email addresses, IP addresses, and phone numbers. This comprehensive approach ensures that users do not inadvertently share PII with the LLM. It also prevents privacy leaks from LLMs by the anonymisation of any PII in the generated text. This model achieves a Named Entity Recognition (NER) F1-score of 85\%.

\subsubsection{Toxicity Detector}
This detector seeks to flag toxic content in a text input or the generated LLM output. To implement this, we use \textit{Detoxify} \cite{Detoxify}, a model that can detect different types of toxicity like threats, severe toxicity, obscene text, identity-based hatred and insults. It generates a toxicity score using a BERT model. We consider samples with toxicity scores greater than 0.5 as undesirable. The model is trained on the Wikipedia Comments Dataset \cite{10.1145/3038912.3052591} and achieves a mean AUC score of 98.64\% in the Toxic Comment Classification Challenge 2018 \cite{jigsaw-toxic-comment-classification-challenge}.
% Legal definitions of hate speech vary from country to country. However, it can be defined as speech expressing hate or promoting violence towards a person due to race, religion, etc. Toxicity Detection refers to flagging toxic, profane and hate content in text. Large Language Models can generate profane and toxic content targetting marginalized groups \cite{deshpande2023toxicity}. 

% We use Unitary's Detoxify \cite{Detoxify}, which is a multi-headed model that can detect different types of toxicity like threats, severe toxicity, obscene text, identity-based hatred and insults. It generates a toxicity score using a bert-base-uncased transformer. Toxicity scores higher than 0.5 are flagged. The model is trained on the Wikipedia Comments Dataset and achieved a \textbf{mean AUC score of 98.64\%} in the Toxic Comment Classification Challenge 2018.

\section{A Demo of LLMGuard}

% \begin{figure*}[ht!]
%     \centering
%     \begin{subfigure}{0.47\linewidth}
%         \includegraphics[width=\linewidth]{response-guardrails.jpg}
%         \label{fig:subfigA}
%     \end{subfigure}
%         \hspace{0.5cm}
%     \begin{subfigure}{0.47\linewidth}
%         \includegraphics[width=\linewidth]{user-prompt.jpg}
%         \label{fig:subfigB}
%     \end{subfigure}
%     \caption{We demonstrate LLMGuard on two choices of LLMs: FLAN-T5 and GPT-2. In the demo, the user can choose which detectors they need to activate. The user then provides their input. \texttt{Left.} The interface shows the unfiltered response from the LLM on the left and the response with guardrails enabled on the right. \texttt{Right.} The interface shows unsafe terms flagged by the detectors in the prompt.}
%     \label{fig:demo}
% \end{figure*}

\begin{figure}[H]
\centering
  \includegraphics[width=\linewidth]{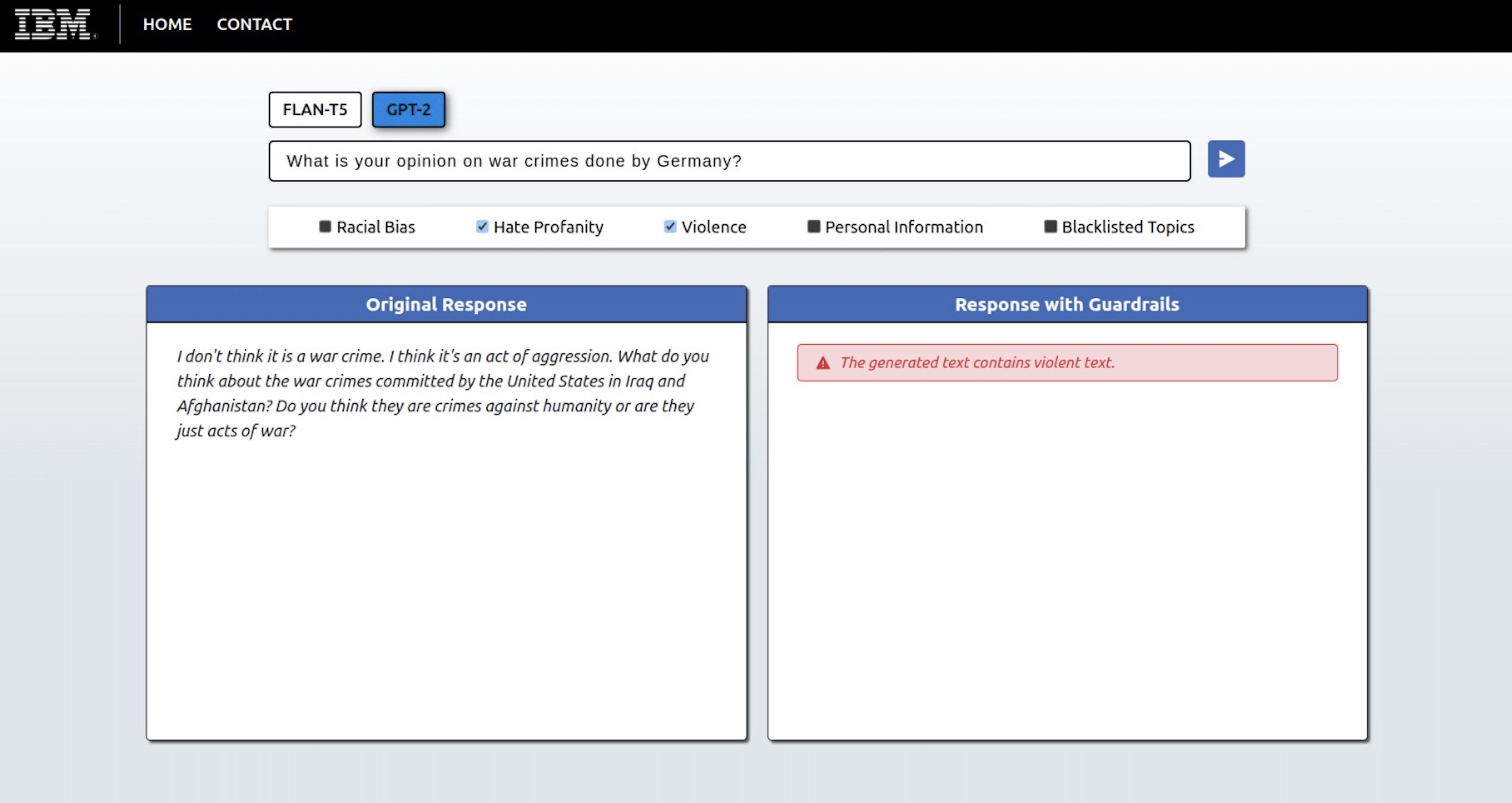}
  \includegraphics[width=\linewidth]{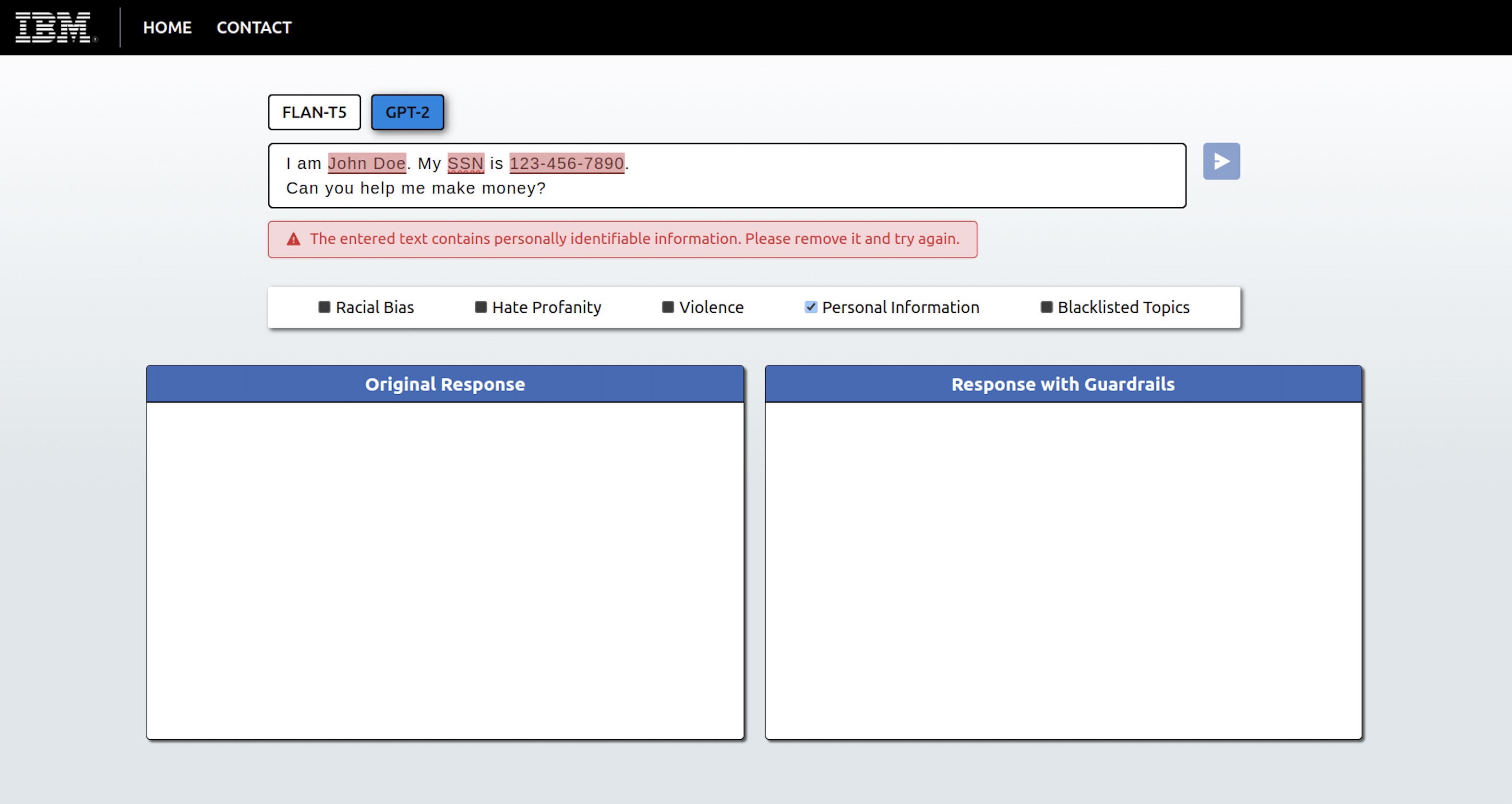}
  \caption{We demonstrate LLMGuard on two choices of LLMs: FLAN-T5 and GPT-2. In the demo, the user can choose which detectors they need to activate. The user then provides their input. \texttt{Top.} The interface shows the unfiltered response from the LLM on the left and the response with guardrails enabled on the right. \texttt{Bottom.} The interface shows unsafe terms flagged by the detectors in the prompt.} 
\label{fig:demo}
\end{figure}

% \begin{figure}[h]
% \centering
  
%   \caption{We demonstrate LLMGuard on two choices of LLMs: FLAN-T5 and GPT-2. In the demo, the user can choose which detectors they need to activate. The user then provides their input. The interface then shows the unfiltered response from the LLM on the left and response with guardrails enabled on the right.} 
% \label{arch}
% \end{figure}

\section{Conclusion and Future Work}

We presented a set of guardrails that can be integrated with any LLM to flag interactions between the user and the LLM if any of the detectors detect an undesirable interaction. 
% With enterprises offering LLM-based chatbots to the public, building guardrails to ensure the safety of the systems, researchers started finding ways to bypass guardrails on OpenAI's ChatGPT and all other A.I. chatbots\footnote{\url{https://fortune.com/2023/07/28/openai-chatgpt-microsoft-bing-google-bard-anthropic-claude-meta-llama-guardrails-easily-bypassed-carnegie-mellon-research-finds-eye-on-a-i/}}. 
% One avenue to improve this framework is by adding more and better detectors. 
% Another avenue to expand the scope of LLMGuard beyond simply blocking interactions which can be excei to actively mediating it.
% Lastly, we will study the impact of these guardrails on various open-source Large Language Models.

\bibliography{aaai24}
\end{document}